\documentclass[9pt, conference]{IEEEtran}
\IEEEoverridecommandlockouts

\usepackage{cite}
\usepackage{amsmath,amssymb,amsfonts}
\usepackage{algorithmic}
\usepackage{graphicx}
\usepackage{textcomp}
\def\BibTeX{{\rm B\kern-.05em{\sc i\kern-.025em b}\kern-.08em
    T\kern-.1667em\lower.7ex\hbox{E}\kern-.125emX}}

\usepackage{url}
\usepackage{multirow}
\usepackage{amsmath}
\usepackage{amssymb}
\usepackage{graphicx}
\usepackage[table,xcdraw]{xcolor}
\usepackage{multirow}
\usepackage{booktabs}
\usepackage{tabularx}
\usepackage{colortbl}

\begin{document}

\title{Multi-Context Temporal Consistent Modeling for Referring Video Object Segmentation \\
}

\author{\IEEEauthorblockN{1\textsuperscript{st} Sun-Hyuk Choi}
\IEEEauthorblockA{\textit{Department of Artificial Intelligence} \\
\textit{Korea University}\\
Seoul, Korea \\
s\_h\_choi@korea.ac.kr}
\and 
\IEEEauthorblockN{2\textsuperscript{nd} Hayoung Jo}
\IEEEauthorblockA{\textit{Department of Artificial Intelligence} \\
\textit{Korea University}\\
Seoul, Korea \\
hayoungjo@korea.ac.kr}
\and
\IEEEauthorblockN{3\textsuperscript{rd} Seong-Whan Lee\textsuperscript{*}} 
\IEEEauthorblockA{\textit{Department of Artificial Intelligence} \\
\textit{Korea University}\\
Seoul, Korea \\
sw.lee@korea.ac.kr}
}

\maketitle
\renewcommand{\thefootnote}{\fnsymbol{footnote}}
\footnotetext[1]{Corresponding author}

\begin{abstract}
Referring video object segmentation aims to segment objects within a video corresponding to a given text description. Existing transformer-based temporal modeling approaches face challenges related to query inconsistency and the limited consideration of context. Query inconsistency produces unstable masks of different objects in the middle of the video. The limited consideration of context leads to the segmentation of incorrect objects by failing to adequately account for the relationship between the given text and instances. To address these issues, we propose the Multi-context Temporal Consistency Module (MTCM), which consists of an Aligner and a Multi-Context Enhancer (MCE). The Aligner removes noise from queries and aligns them to achieve query consistency. The MCE predicts text-relevant queries by considering multi-context. We applied MTCM to four different models, increasing performance across all of them, particularly achieving 47.6 $\mathcal{J\&F}$ on the MeViS. Code is available at \textit{\url{https://github.com/Choi58/MTCM}}.
\end{abstract}

\section{introduction}

With the advancement of artificial intelligence, various fields such as computer vision \cite{hwang2006full, lee1999integrated, roh2007accurate, lee1990translation} and signal processing \cite{dhar2023glgan, mane2020multi, jeong2019classification} have been gaining attention. Referring Video Object Segmentation (RVOS) is one of the vision-language learning \cite{zhang2024fine, wang2024text, le2024waver, feng2024crestyler}, where the goal is to identify and segment objects in a video corresponding to a given text description. RVOS is challenging because it involves identifying the object corresponding to the text at the pixel level within each frame while also leveraging information from other frames to accurately locate the target. Therefore, RVOS models require the integration of understanding across different modalities in each frame as well as the relationships between multiple frames.

In the early stages of RVOS, methods including dynamic convolution \cite{gavrilyuk2018actor, wang2020context} and cross-attention mechanisms \cite{khoreva2019video, wang2019asymmetric, seo2020urvos, ye2021referring, chen2021cascade, hui2021collaborative, wu2022multi, ning2020polar} were commonly used. To simplify the pipeline and improve efficiency, \cite{botach2022end, wu2022language} proposed transformer-based approaches. These methods integrate cross-modal interaction with pixel-level understanding, facilitating better alignment between different modalities. However, they fail to consideration of the temporal relationship between frames.

Recently, transformer decoders were introduced into transformer approaches to enhance the correlation between frames. There are two types of decoders: frame-level decoders and video-level decoders. The frame-level decoders \cite{luo2024soc, han2023html, tang2023temporal} aggregate global context for each query and update it. Afterwards, frame-level decoders generate individual mask embeddings for each frame. Since these decoders assume that the same query always targets the same object across frames, the low query consistency could degrade performance if the query indicates a different object in the middle of the video. In the other hand, the video-level decoders \cite{ding2023mevis, he2024decoupling} combine all queries to create video-level queries for global context. Instead of generating individual mask embeddings for each frame, video-level decoders use a single unified mask embedding to generate masks across frames. Because these methods learn the overall instance features throughout the video, they could miss detailed features of individual frames. Furthermore, both frame-level and video-level decoders only focus on global context and overlook relationships between adjacent frames, limiting the ability to capture short-term actions.

\begin{figure}[!t]
    \centering
    \includegraphics[width=\linewidth]{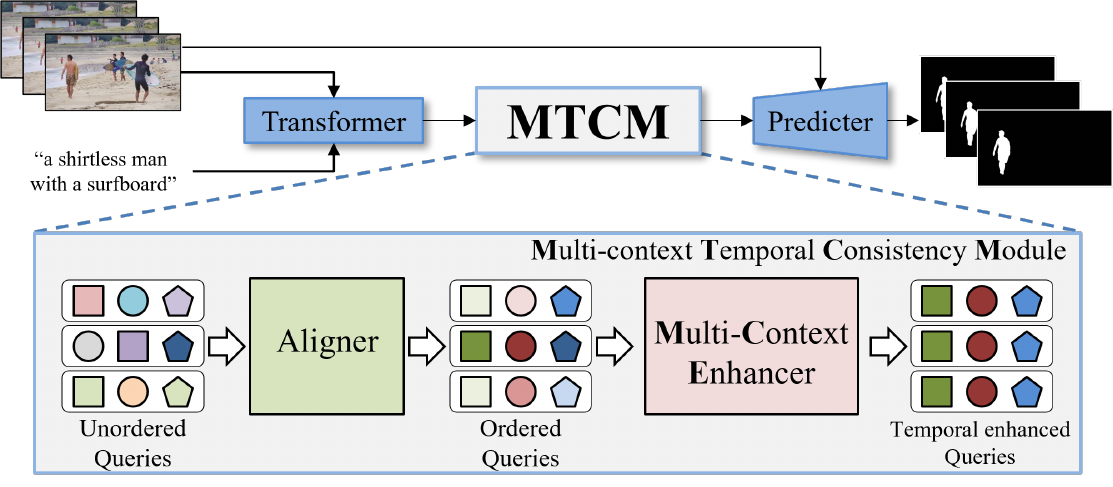}
    \caption{Overview of the proposed module, which consists of an Aligner and a Multi-Context Enhancer (MCE). The Aligner improves query consistency, while the MCE selects objects by considering both local and global contexts.}
    \label{fig:1}
\end{figure}

\begin{figure*}[ht]
    \centering
    \begin{minipage}[b]{0.45\textwidth}
        \centering
        \includegraphics[width=\textwidth]{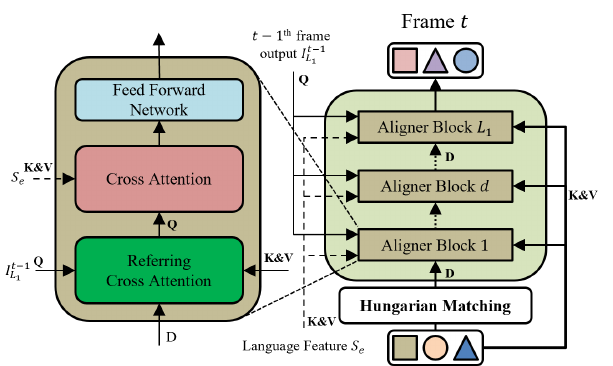} 
        {(a) Aligner}
        \label{fig:aligner}
    \end{minipage}
    \hfill
    \begin{minipage}[b]{0.45\textwidth}
        \centering
        \includegraphics[width=\textwidth]{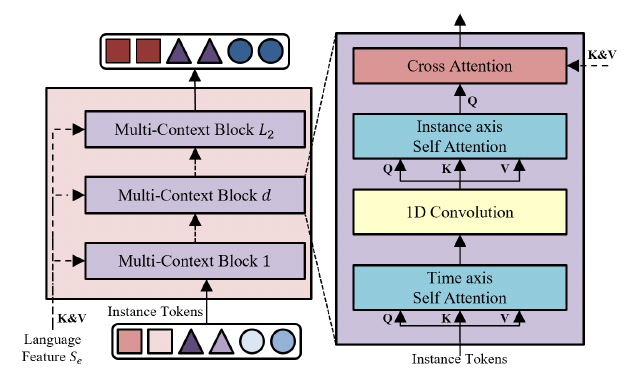} 
        {(b) Multi-Context Enhancer}
        \label{fig:mce}
    \end{minipage}
    \caption{The structure of the Aligner (a) and the Multi-Context Enhancer (b). The Aligner aligns the queries and removes irrelevant information by utilizing queries from the previous frame, ensuring that each query shares common features. The MCE captures the multi-context of each query to supplement the information of each frame, enabling accurate object selection.}
    \label{fig:2}
\end{figure*}

To address the above issues, temporal modeling needs to capture the multi-context of queries with improved query consistency while also providing detailed information for each frame. Therefore, we propose the Multi-context Temporal consistency Module (MTCM), which is applied to transformers as shown in Fig. \ref{fig:1}. MTCM consists of an Aligner and a Multi-Context Enhancer (MCE). The Aligner arranges queries and removes unnecessary information, making it easier to capture temporal context. We refer to this effect as query consistency. The MCE captures and reflects the local and global contexts of the queries to understand the short-term actions and overall movements of the instances. The MCE then compares the temporally enhanced queries with the text to predict the target. We applied MTCM to four different models and achieved performance improvements on three datasets.

Our main contributions are as follows:
\begin{itemize}
    \item We propose the MTCM module, which is applicable to various models with transformer architectures to enhance temporal modeling.
    \item We introduce the Aligner to enhance query consistency for easier understanding the context of each query.
    \item We introduce the MCE to determine target-related instances considering both local and global contexts for the correct selection of targets.
\end{itemize}

\section{Related Works}

\textbf{Referring Video Object Segmentation.} \cite{gavrilyuk2018actor} firstly proposed the task of actor and action video segmentation from a sentence along with the dataset. Subsequently, \cite{khoreva2019video, seo2020urvos} proposed datasets and the RVOS task, which handles unconstrained expressions, to address the limitations of previously restricted expressions. Recently, researches \cite{ding2023mevis, he2024decoupling} have emerged focusing on segmenting objects in more dynamic videos using motion-based expressions. 

Architecture of RVOS are broadly categorized into dynamic convolution, cross-attention, and transformer-based methods. Dynamic convolution-based methods encode text features as kernels to convolve over visual features. \cite{wang2020context} improved it by enhancing spatial information. However, these approaches still face difficulties with natural language variability and global context modeling. 

Cross-attention-based methods perform cross-attention at the pixel level and generate the mask through FCN. Later improvements, such as \cite{hui2021collaborative} and \cite{ding2022lbdt}, addressed spatial misalignment and introduced explicit spatiotemporal interactions, however they still face complex pipelines and limited object-level utilization.

Transformer-based methods adopted end-to-end approaches by integrating a small set of object queries conditioned on language to capture object-level information.\cite{miao2023spectrum} proposed a spectrum-guided multi-granularity method to address feature drift during segmentation. \cite{he2024decoupling} introduced a technique that separates static and motion perception to enhance temporal understanding. However, these approaches often assume that one query corresponds to one object and primarily consider either local or global context, which limits their effectiveness. To address these limitations, we propose the Multi-Context Temporal Consistency Module (MTCM) to improve query consistency and perform temporal modeling with multiple context considerations.

\section{methodology}

\subsection{Overview}
The overview of the proposed module is shown in Fig. \ref{fig:1}. The Aligner receives the instance tokens generated by the transformer as input. Then, the Aligner enhances the query consistency by reordering the instance tokens and removing information that is unrelated to the previous frame. The Multi-Context Enhancer (MCE) emphasizes target-related instances in each frame by considering both local and global contexts. Finally, the predictor generates the masks.

\subsection{Aligner}
\label{sec:aligner}

Instance tokens generated by the transformer indicate different instances across frames even for the same query. The Aligner reorders the instance tokens per frame to ensure they refer to the same instance. As shown in Fig. \ref{fig:2} (a), the Aligner includes $L_1$ blocks, with each block consisting of a Referring Cross Attention layer (RCA) \cite{zhang2023dvis}, a Cross-Attention layer (CA), and a Feed-Forward Neural Network (FFN). 

The Aligner takes the instance tokens $ \{ O^t \mid t \in [1, T], O^t \in \mathbb{R}^{N \times C} \}$, which are $N$ candidate instance tokens generated by the transformer for each frame. $T$ and $C$ denote the length of the video and the number of channels. First, hungarian matching \cite{kuhn1955hungarian} orders the instance tokens using cosine similarity as the cost. This process can be formulated as: 
\begin{equation}
\begin{cases}
\tilde{O}^t = \text{Hungarian}(\tilde{O}^{t-1}, O^t), & t \in [2, T] \\
\tilde{O}^t = O^t, & t = 1,
\end{cases}
\label{eq:hungarian matching}
\end{equation}
where $\tilde{O}^t$ is the aligned instance tokens that still contain unnecessary information.

Secondly, to ensure that each query has identical information, the RCA layer denoises the aligned instance tokens by utilizing information from the previous frame. This process can be formulated as:
\begin{equation}
\dot{I}^{t}_{d} = \text{RCA}(I^{t}_{d-1}, I^{t-1}_{L_1}, \tilde{O}^t, \tilde{O}^t),
\label{eq:rca2}
\end{equation}
\begin{equation}
\text{RCA}(D, Q, K, V) = D + \text{MHA}(Q, K, V),
\label{eq:rca1}
\end{equation}
where, \text{MHA} stands for multi-head attention \cite{vaswani2017attention}. \(D\), \(Q\), \(K\), and \(V\) denote the residual information, query, key, and value. $t$ and $d$ represent the index of the time, and the index of the layer. $I^{t-1}_{L}$ refers to the output of the previous frame. Because the processes above remove target instance information, the cross-attention layer reintroduces target information to the token through language features. Finally, the FFN outputs aligned and denoised instance tokens for the target frame. This process can be formulated as:
\begin{equation}
I^{t}_{d} = \text{FFN} ( \text{CA}(\dot{I}^{t}_{d}, S_e) ),
\label{eq:CA + FFN}
\end{equation}
where CA uses the first argument as the query and uses the second argument as the key and value. $S_e$ denotes language features.
Through the processes above, the same instance queries point to the same object across different frames and share similar features. The enhanced query consistency facilitates in understanding the context of each instance.

\subsection{Multi-Context Enhancer} 
\label{sec:enhancer}
Since the text corresponds to a part or all of the video, the MCE is used to compare partial or entire context of each object with the text to determine how closely each object is related to the text. Additionally, the MCE selects which object is relatively closer to the target among similar objects to make the selection. As shown in Fig. \ref{fig:2} (b), MCE includes $L_2$ blocks, with each block consisting of a Time axis Self-Attention layer (TSA), a 1D convolution layer (Conv), an Instance axis Self-Attention layer (ISA), and a Cross-Attention layer (CA). 

In the time axis self-attention layer, self-attention is performed on the tokens of the Aligner $I$ along the time axis to consider the overall context. To enhance the local context of each query, they pass through a 1D convolution layer. This process can be formulated as:
\begin{equation}
\dot{Q} = \text{Conv}({\text{TSA}(I)}). 
\label{eq:CA2}
\end{equation}

In the instance axis self-attention layer, self-attention is performed along the instance axis to understand the relative relationships among the queries. Finally, a cross-attention layer determines which query is the target among the temporally enhanced queries. This process can be formulated as:
\begin{equation}
\hat{Q} = \text{CA}( \text{ISA}( \dot{Q}), S_e).
\label{eq:CA3}
\end{equation}

The instance tokens which are refined by MCE recognize the target through the multi context of the video and maintain temporal consistency.
\subsection{Module-wise Training Strategy} 
We train the entire framework in the order of transformer, the Aligner, and the MCE to suppress noise during training and allow each module to focus on its specific role. In the training stage of the Aligner, once the instance features are properly generated, it becomes easier to refine the characteristics of each query and improve consistency. In the training stage of the MCE, the removal of noise and the uniformity of features across queries help in understanding multiple contexts. The training process is as follows: first, we train the transformer, then freeze the transformer and train the Aligner, Finally, we freeze the other modules and train the MCE. The process above allows each module to focus on its specific stage.


\begin{figure*}[htbp]
    \centering
    \begin{minipage}[b]{\textwidth}
        \centering
        \includegraphics[width=\textwidth]{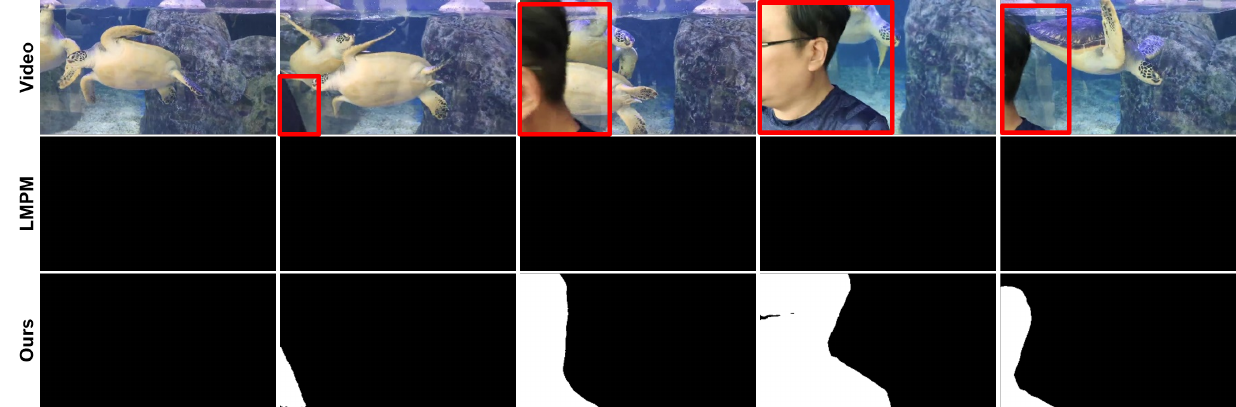}
        {(a) ``The individual observing the turtles in the enclosure.''}\label{fig:3a}
    \end{minipage}
    \begin{minipage}[b]{\textwidth}
        \centering
        \includegraphics[width=\textwidth]{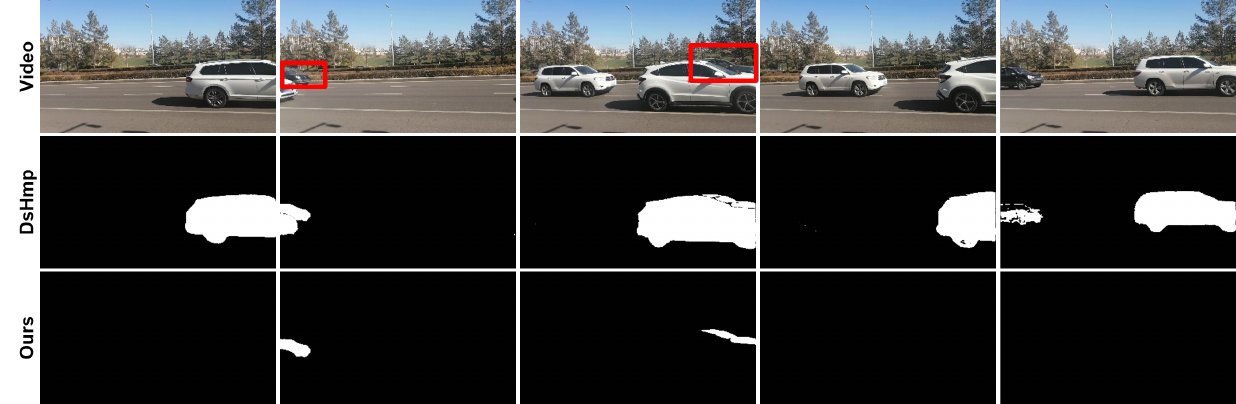}
        {(b) ``The initial dark car moving to the right.''}\label{fig:3d}
    \end{minipage}
    
    \caption{Qualitative comparison of our method with LMPM and DsHmp. Red boxes indicate the targets. (a) and (b) are the given text queries respectively. Both videos are challenging samples where the object is observed in the middle of the video.}
    \label{fig:3}
\end{figure*}

\section{experiments}

\subsection{Experiment Setup}
\textbf{Datasets and metrics.} We evaluated our method on three datasets: MeViS \cite{ding2023mevis}, A2D Sentences \cite{gavrilyuk2018actor}, and JHMDB Sentences \cite{gavrilyuk2018actor}. MeViS is a newly proposed dataset focused on dynamic information. A2D Sentences is a dataset for actor and action segmentation. JHMDB Sentences is a dataset that contains 21 different actions. For evaluation metrics, MeViS uses $\mathcal{J}$, $\mathcal{F}$, and $\mathcal{J\&F}$, while A2D-S and JHMDB-S use IoU, following \cite{ding2023mevis} and \cite{gavrilyuk2018actor}, respectively.


\textbf{Baseline.} We applied the proposed method to ReferFormer \cite{wu2022language}, SgMg \cite{miao2023spectrum}, LMPM \cite{ding2023mevis} and DsHmp \cite{he2024decoupling}. Both ReferFormer and SgMg use the Video Swin Transformer \cite{liu2022video}, and RoBERTa \cite{liu2019roberta} as their backbone, and employ Deformable DETR \cite{deformable_detr} as the transformer. LMPM and DsHmp both use the Swin Transformer \cite{liu2021swin} and RoBERTa as their backbone, and employ Mask2Former \cite{cheng2022mask2former} as the transformer. After removing the motion decoder, the MTCM is applied.

\subsection{Implementation Details} 
We first trained each baseline according to the method proposed by that baseline, then sequentially trained the Aligner and the MCE using their settings. Across all frameworks, the Aligner and the MCE were configured with 6 layers and trained with a batch size of 2 for 40,000 iterations. For ReferFormer and SgMg, both the Aligner and the MCE were trained using 5 frames. For LMPM, we used 5 frames for the Aligner and 21 frames for the MCE. For DsHmp, both were trained with 8 frames.
\subsection{Results}
\begin{table}[]
\centering
\begin{center}
\caption{Quantitative Comparison with Other Methods on MeViS. The Relatively Better Results are Highlighted in Bold.}
\begin{tabularx}{\columnwidth}{c|>{\centering\arraybackslash}X >{\centering\arraybackslash}X >{\centering\arraybackslash}X}

\toprule[1.5pt]
                         & \multicolumn{3}{c}{MeViS} \\
\multirow{-2}{*}{Method} & $\mathcal{J\&F}$ & $\mathcal{J}$ & $\mathcal{F}$ \\ \hline
URVOS \cite{seo2020urvos}& 27.8    & 25.7   & 29.9   \\
LBDT \cite{ding2022lbdt} & 29.3    & 27.8   & 30.8   \\
MTTR \cite{botach2022end}& 30.0    & 28.8   & 31.2   \\
ReferFormer \cite{wu2022language} & 31.0    & 29.8   & 32.2   \\
VLT + TC \cite{ding2022vlt} & 35.5    & 33.6   & 37.3   \\ \hline
LMPM \cite{ding2023mevis}  & 37.2    & 34.2   & 40.2   \\ 
LMPM + MTCM               & \textbf{42.3}    & \textbf{38.4}   & \textbf{46.4}   \\
DsHmp \cite{he2024decoupling} & 46.4    & 43.0   & 49.8   \\
DsHmp + MTCM              & \textbf{47.6} & \textbf{44.1}   & \textbf{51.1}   \\ \bottomrule[1.5pt]
\end{tabularx}%
\label{mevis}
\end{center}
\end{table}

\begin{table}[]
\centering
\caption{Quantitative Comparison with Other Methods on A2D-S and JHMDB-S. The Relatively Better Results are Highlighted in Bold.}
\label{other datasets}
\begin{tabularx}{\columnwidth}{c|>{\centering\arraybackslash}X >{\centering\arraybackslash}X|>{\centering\arraybackslash}X >{\centering\arraybackslash}X}

\toprule[1.5pt]
                         & \multicolumn{2}{c|}{A2D-S}      & \multicolumn{2}{c}{JHMDB-S}    \\
\multirow{-2}{*}{Method} & oIoU          & mIoU          & oIoU          & mIoU          \\ \hline
Gavrilyuk \textit{et al.} \cite{gavrilyuk2018actor} & 53.6          & 42.1          & 54.1          & 54.2          \\
ACGA \cite{wang2019asymmetric} & 60.1          & 49.0          & 57.6          & 58.4          \\
CSTM \cite{hui2021collaborative} & 66.2          & 56.1          & 59.8          & 60.4          \\
MTTR \cite{botach2022end} & 72.0          & 64.0          & 70.1          & 69.8          \\
HTML \cite{han2023html}   & 77.6          & 69.2          & -             & -             \\
SOC \cite{luo2024soc}     & 78.3          & 70.6          & 72.7          & 71.6          \\ \hline
ReferFormer \cite{wu2022language} & 77.6          & 69.9          & 71.9          & 71.0          \\
ReferFormer + MTCM        & \textbf{78.0} & 69.9 & \textbf{72.0}  & 71.0          \\
SgMg \cite{miao2023spectrum} & 78.0          & 70.4          & 72.8          & 71.7 \\
SgMg + MTCM               & \textbf{78.7} & \textbf{70.7} & \textbf{72.9} & 71.7 \\ \bottomrule[1.5pt]
\end{tabularx}%
\end{table}

\textbf{Quantitative results.} As shown in Table. \ref{mevis} and \ref{other datasets}, applying MTCM to the four models led to performance improvements. In MeViS, the $\mathcal{J\&F}$ scores increased by $+5.1$ and $+1.2$ for LMPM and DsHmp, respectively. In A2D Sentences and JHMDB Sentences, the oIoU improved modestly by $+0.4$ and $+0.7$ for ReferFormer and SgMg, respectively. For datasets like JHMDB-S and A2D-S with few annotations per video, it is challenging to learn temporal continuity, resulting in relatively smaller performance improvements for our model.

\textbf{Qualitative results.} As shown in Fig. \ref{fig:3}, our model consistently tracks the target object across various frameworks when the text-related content or the object itself came out in the middle of the video. In Fig. \ref{fig:3} (a), the man appears partially, making segmentation difficult. Our method effectively segmented the object using temporal information, while LMPM failed to segment anything due to low confidence. In Fig. \ref{fig:3} (b), without focusing on ``initial'', it becomes a challenging sample to distinguish the dark cars. Our method accurately focused on ``initial'' and did not track the second dark car, while DsHmp segmented other objects.


\subsection{Ablation Study} 
We conducted experiments on DsHmp to evaluate the effects of each module and training strategy on MeViS validation set. 

\textbf{Modules.} As shown in Table. \ref{tab:all ablation study}, both modules contributed to improvements in performance. Especially, the Aligner contributed an average improvement of +4.9 in $\mathcal{J\&F}$ scores. The MCE increased the $\mathcal{J\&F}$ scores by an average of +8.3. It was observed that the MCE contributed relatively more to performance improvements compared to the Aligner. The Aligner focuses on noise-reducing alignment algorithms, but in datasets like MeViS, which involve various objects, it lacks the ability to highlight targets. Therefore, the multi-context perspective of MCE appears to be more effective.

\begin{table}[]
\centering
\caption{Ablation Studies of Aligner, MCE, and Module-wise Training Strategy. The Best Results are Highlighted in Bold.}
\label{tab:all ablation study}
\begin{tabularx}{\columnwidth}{>{\centering\arraybackslash}X >{\centering\arraybackslash}X >{\centering\arraybackslash}X >{\centering\arraybackslash}X >{\centering\arraybackslash}X >{\centering\arraybackslash}X}

\toprule[1.5pt]
Aliger & MCE & Strategy & $\mathcal{J\&F}$ & $\mathcal{J}$ & $\mathcal{F}$    \\ \hline
  &   &                                 & 46.9 & 42.3 & 51.5 \\
\checkmark &   &                        & 51.1 & 46.9 & 55.1  \\
  & \checkmark &                        & 54.8 & 49.8 & 59.8 \\
\checkmark & \checkmark &               & 49.4 & 44.3 & 54.4 \\ \hline
\checkmark &   & \checkmark             & 52.6 & 48.2 & 57.0 \\
  & \checkmark & \checkmark             & 55.7 & 51.1 & 60.3 \\
\rowcolor[HTML]{EFEFEF}
\checkmark & \checkmark & \checkmark    & \textbf{56.1} & \textbf{51.6} & \textbf{60.6} \\ \bottomrule[1.5pt]
\end{tabularx}%
\end{table}

\textbf{Module-wise training strategy.} According to Table. \ref{tab:all ablation study}, applying the training strategy to each module improved performance, while performance significantly decreased without the strategy. It shows that the learned queries greatly assist in the alignment performed by the Aligner and help distinguish the features between queries and highlight the target in the MCE.


\section{conclusion}
We propose MTCM which enhances the temporal modeling of the transformer model. MTCM includes an Aligner to improve query consistency and an MCE to enrich frame information by considering both local and global contexts. We use a training strategy suitable for the proposed method. The proposed method is successfully applied to various models, increasing performance.

\section*{Acknowledgment}
This work was supported by the Institute for Information \& communications Technology Planning \& Evaluation(IITP) grant funded by the Korea government(MSIT) (No. RS-2019-II190079, Artificial Intelligence Graduate School Program(Korea University)) and Center for Applied Research in Artificial Intelligence (CARAI) grant funded by DAPA and ADD (UD230017TD).

\bibliographystyle{IEEEtran}
\bibliography{IEEEabrv,mybibfile}

\end{document}